# 3D Adapted Random Forest Vision (3DARFV) for Untangling Heterogeneous-Fabric Exceeding Deep Learning Semantic Segmentation Efficiency at the Utmost Accuracy

Omar Alfarisi, Zeyar Aung, Qingfeng Huang, Ashraf Al-Khateeb, Hamed Alhashmi, Mohamed Abdelsalam, Salem Alzaabi, Haifa Alyazeedi, and Anthony Tzes

## 1. Abstract

Planetary exploration depends heavily on 3D image data to characterize the static and dynamic properties of the rock and environment. Analyzing 3D images requires many computations, causing efficiency to suffer lengthy processing time alongside large energy consumption. High-Performance Computing (HPC) provides apparent efficiency at the expense of energy consumption. However, for remote explorations, the conveyed surveillance and the robotized sensing need faster data analysis with ultimate accuracy to make real-time decisions. In such environments, access to HPC and energy is limited. Therefore, we realize that reducing the number of computations to optimal and maintaining the desired accuracy leads to higher efficiency. This paper demonstrates the semantic segmentation capability of a probabilistic decision tree algorithm, 3D Adapted Random Forest Vision (3DARFV), exceeding deep learning algorithm efficiency at the utmost accuracy.

## 2. Introduction

Machine learning algorithms [1-5] advanced analyzing geo-sensors data [6] for delivering quality real-time decisions. Determining heterogeneous porous media (e.g., Cretaceous carbonate) physical and chemical properties is vital to decipher, characterize and classify its elements for building representative 3D geological models, flow simulation, and future development decisions [7-15]. However, for the planetary exploration [16], some decisions are needed fast enough not to miss an opportunity or suffer costly challenges [17-19]. Micro-nano-porous media images determined heterogeneous fabric physical properties (e.g., porosity, pore throat network, permeability, capillary pressure, wettability, relative permeability) [20-29]. Computer vision also determined heterogeneous fabric chemical properties (e.g., lithology and mineral volume) [30, 31], leading to digital rock typing (DRT), an artificial intelligence geoscience research agent. Although the advances in computation power, including HPC and quantum computing, provide faster computation, HPC is not available or accessible at every remote Earth, Moon, or Mars location. Even with HPC, developing an algorithm with several folds' efficiency is more environmentally friendly with less power consumption. Therefore, identifying algorithms with higher efficiency at optimal accuracy is highly attractive for scholars and industrial applications [32]. Thus, our research addressed higher efficiency at ultimate accuracy for supervised semantic segmentation through four stages of algorithms elimination. We investigate the most widely used machine learning algorithms, k-Nearest Neighbour (kNN), Logistic Regression (LoR or LR), Naive Bayes (NB), Support Vector Machine (SVM), Random Forest (RF) [32], Multi-Layer Perceptron (MLP) [3, 33-37], Deep Neural Network (DNN) [1-4], and Convolutional Neural Network (CNN) [2, 38]. We increased the data complexity at each stage for the four elimination stages, and with it, the computation time increased. Two finalist algorithms reached the ultimate accuracy of 1.00, Deep Neural Network and Random Forest, at which the latest delivered the highest efficiency for supervised semantic segmentation of heterogeneous fabric.

## 3. Method

We propose investigating several machine learning algorithms on multiple stages and eliminating lower accuracy and efficiency. Neural network-based algorithms, including deep neural network (DNN), also called Deep Learning, have the highest accuracy and lowest efficiency [39]. Therefore, we target identifying Non-Neural-Network-based-Algorithms (NNNA), generally characterized by highest efficiency and lower accuracy [32] than NN algorithms. We compare algorithms by maintaining the same computation power (CPU and GPU) in all stages. The parameters and hyperparameters are the default ones recommended by the software libraries, while for Random Forest, we used the recommended Adapted Difference of Gaussian Random Forest algorithm [31].

### 3.1- 3DARFV First Stage: Evaluating Algorithms' Accuracy and Efficiency for Artificial Data

In the first stage, we consider several NNNA and omit NN algorithms [32] because this stage aims to identify the most accurate NNNA algorithm that potentially reaches DNN accuracy. We use artificial data for this stage only, using domain expert knowledge to generate heterogeneous morphology-based synthetic data [24, 32], as shown in Table 1. We discuss the findings of this stage in the Results and Discussions section of this paper.

Table 1. Artificial data sample mimics domain expert, geoscientist (petrophysicist) knowledge [32].

| | PhiXSectContin | PixelColor | NeighbColorGrad | Betw2Amplify | Lable |
|---|---|---|---|---|---|
| 0 | 0 | 251 | 64 | 0 | Solid |
| 1 | 1 | 78 | 19 | 1 | thraot |
| 2 | 0 | 138 | 29 | 0 | NC_Vugs |
| 3 | 0 | 133 | 35 | 1 | NC_Vugs |
| 4 | 1 | 185 | 45 | 0 | Solid |
| 5 | 1 | 96 | 84 | 0 | Pore |

### 3.2- 3DARFV Second Stage: Evaluating Algorithms' Accuracy and Efficiency for MNIST

In the second stage, we studied the performance of the algorithms for black and white image data, MNIST data set with ten classes as described in Table 2 and Fig. 1. We use NN-based algorithms in this stage, DNN, Deep CNN (two filter layers), and Recurrent Neural Network (RNN) algorithms. This stage aims to compare the accuracy and efficiency of multiple NN-based algorithms.

Table 2. MNIST data set splits to training, validation, and test data sets.

| | |
|---|---|
| Dataset | Name: mnist |
| | Type: public |
| | Samples: 70000 |
| Samples | training: 56000 |
| | validation: 7000 |
| | test: 7000 |
| | split: 3 |

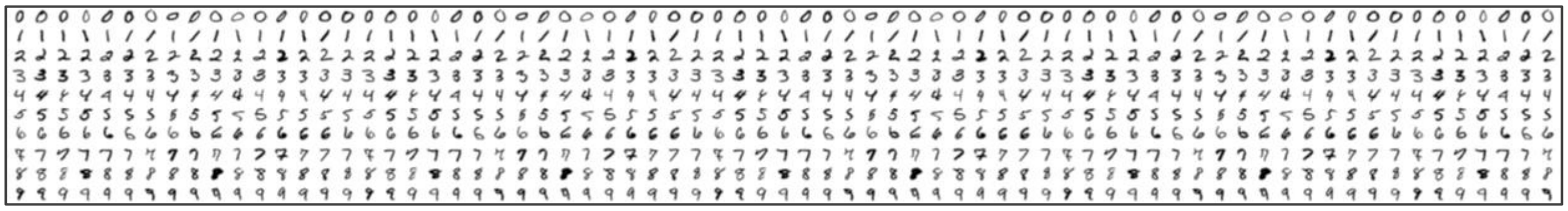

Fig. 1. MNIST data set sample (images of labeled single-digit handwritten Arabic numbers), used for the second stage of this research of choosing an accurate-efficient algorithm, contains back and white photos of 10 classes.

### 3.3- 3DARFV Third Stage: Evaluating Algorithms' Accuracy and Efficiency for CIFAR-10

In the third stage, we increased the complexity of the experiment with colored image data set of ten classes, the CIFAR data set. In this stage, we made two changes: introducing color to the images and adding objects while maintaining the same number of classes as described in Table 3 and Fig. 2. We use NN-based algorithms in this stage, Deep CNN (two filter layers) and Deeper CNN (4 filter layers). This stage compares the accuracy and efficiency of multiple filter layers effects on CNN-based algorithms.

Table 3. CIFAR-10 data set split into training and validation data sets.

| | |
|---|---|
| Dataset | Name: cifar-10 |
| | Type: public |
| | Samples: 60000 |
| Samples | |
| | training: 48000 |
| | validation: 12000 |
| | test: 0 |
| | split: 1 |

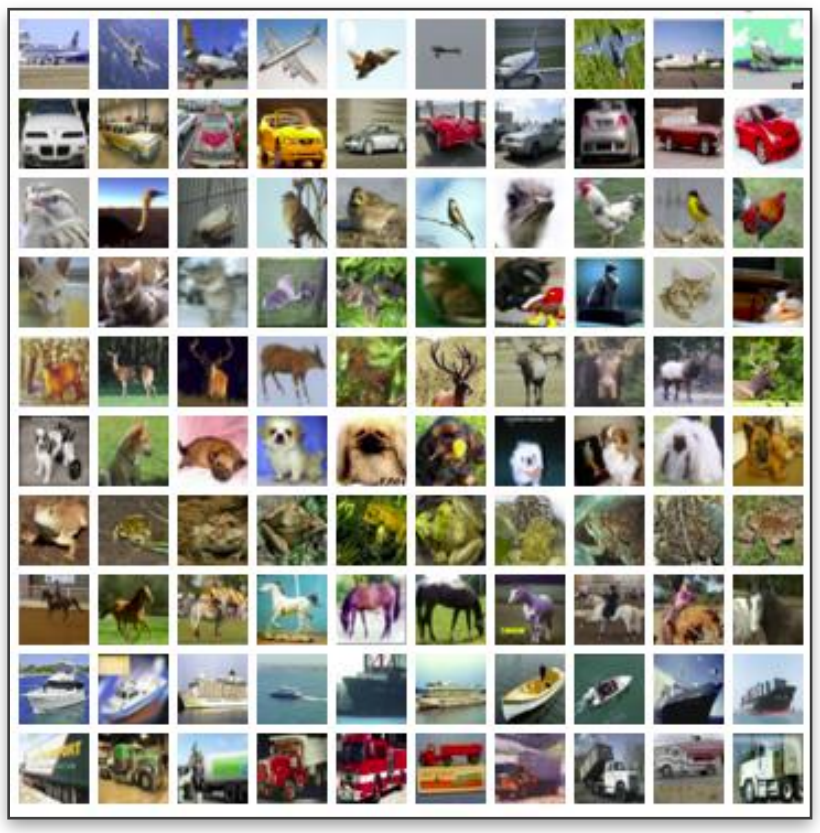

Fig. 2. CIFAR-10 data set sample (images of labeled one of 10 mutually exclusive classes: airplane, car, bird, cat, deer, dog, frog, horse, ship, and truck) contains color images of 10 classes.

### 3.4- 3DARFV Fourth Stage: Evaluating Algorithms' Accuracy and Efficiency for Natural Heterogeneous Carbonate Rock

In the fourth stage, we use all the learning from the first three data sets to experiment with the fourth data set, Savonnières Carbonate (SAV-II). It is gray-scale X-Ray micro-Computerized Tomography (μCT) images of natural heterogeneous carbonate rock. Planetary exploration missions in Earth, Moon, and Mars would use μCT for rock properties characterization. This data set has 256 gray shads, which we labeled five classes representing physical (i.e., pore, grain size, bioclast, etc.) and chemical properties (i.e., Pyrite, Calcite, Dolomite, etc.). Domain expert knowledge is vital for defining these classes because, for non-domain experts, an apparent overlap between classes masks the physical representation. Unsupervised learning approaches for such data would suffer the exact miss-representation due to proximity between semantic segmentation features clusters.

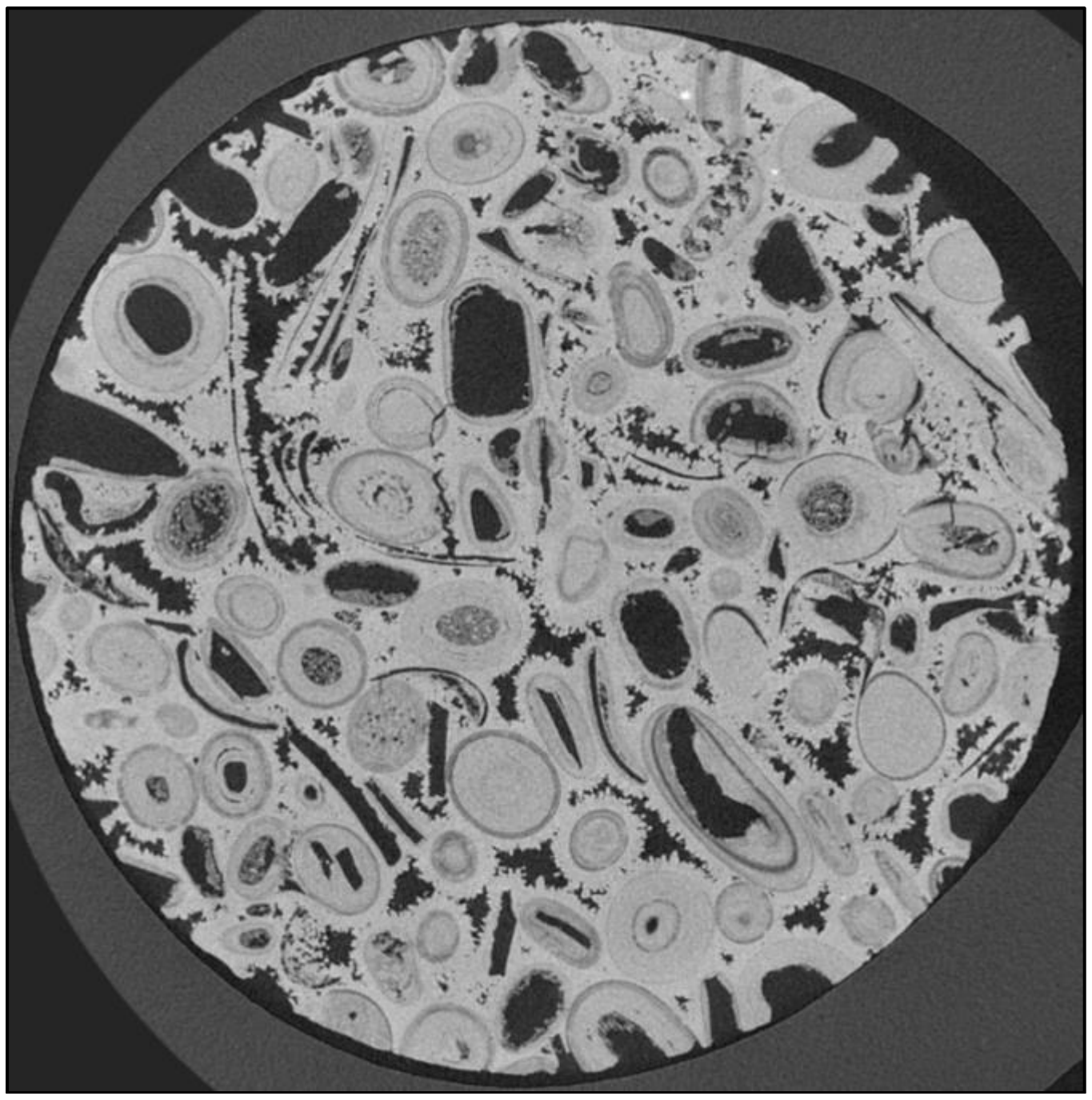

Fig. 3. Savonnières Carbonate (SAV-II), Natural Cretaceous Rock μCT image [40].

We identify five classes in the μCT image of the Savonnières carbonate (SAV-II) [40]. The carbonate rock has a heterogeneous morphology due to marine biological deposits. Fig. 4 shows μCT images labels of five pore-throat size networks (PorThN): i- Isolated Bioclast (Green-label), ii- Carbonate Cement (Yellow-label), iii- Intragranular Connected Vugs and Mini-Fractures (Purple-label), iv- Heavy Mineral of Pyrite (Blue-label), and v- Intergranular Pore (void). The labeling process is efficient because it takes a few minutes only to label all the five classes using open source software and plugin, Fiji [41] and Weka [35, 36].

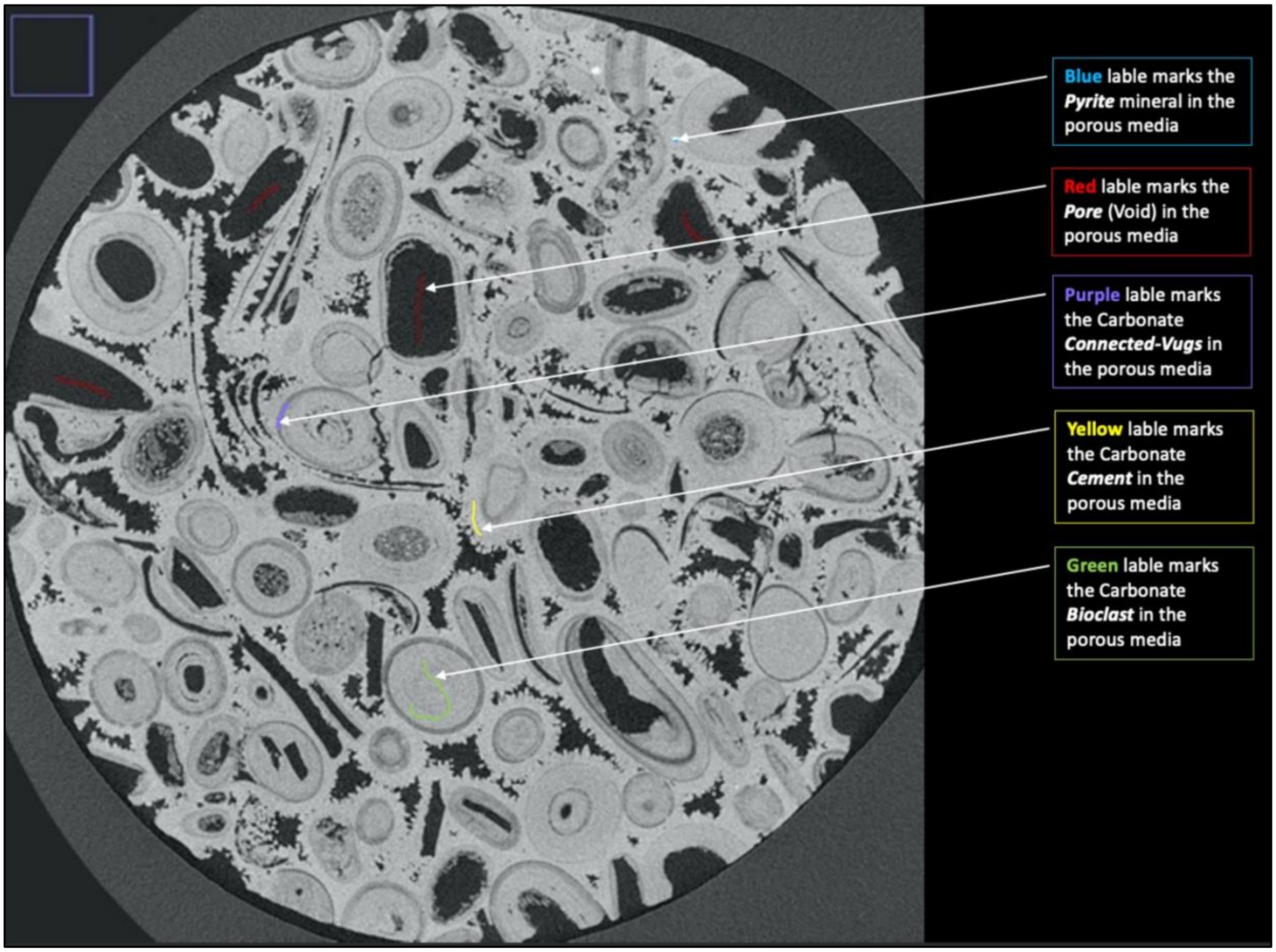

Fig. 4. Labels of Pore Throat Size Networks (PorThN) of Savonnières Cretaceous carbonate (SAVII) rock µCT images.

We use the qualified NN-based algorithm, DNN, and three NNNA algorithms, NB, SVM, and RF, in this stage. In addition, we added Shallow NN (also called Multi-Layer Perceptron (MLP)) because MLP consumes less processing power and time than DNN. However, MLP has less accuracy than DNN in general. This stage aims to compare the efficiency of NN and NNNA algorithms.

## 4. Results and Discussions

### 4.1- First Stage Results and Discussions for analyzing Artificial Data

We run the NNNA algorithms with similar features and the recommended default parameters. In Table 4, we display the results of NNNA accuracy. We did not consider computation time for elimination at this stage. We target the accurate NNNA algorithms to be qualified for the subsequent steps to compete with NN-based algorithms. We conclude that kNN and RF hold higher accuracy than LoR, NB, and SVM. We also notice that LoR has the lowest accuracy among NNNA. Therefore, we exclude LoR from subsequent stages. We also conclude that NNNA can deliver ultimate accuracy for structured data. We recommend parameter tuning for all algorithms for targeting higher accuracy and shorter computation time for NNNA, using automated machine learning [42] (AutoML) engines [43]. AutoML takes more extended computation because it combines the tuning of algorithms, parameters, and hyperparameters that aggregates all the time needed for individual algorithms [44]. However, we recommend AutoML for all subsequent stages, but after going through these initial elimination stages (Four Stages) to reduce the AutoML computation time. Hence, the results of the algorithms comparison of this paper provide a heads up for researchers in selecting the algorithms that best suit the different data sets based on computation time and accuracy.

Table 4. The Accuracy Results of the First Stage compare the Non-Neural-Network-based-Algorithms (NNNA).

| | | Stage 1 |
|---|---|---|
| # | Algorithm | Artificial Data - Accuracy |
| 1 | LoR - Logistic regression | 0.9100 |
| 2 | NB - Naive Bayes | 0.9300 |
| 3 | SVM - Support Vector Machine | 0.8500 |
| 4 | KNN | 1.0000 |
| 5 | Shallow NN - MLP - Multi Layer Perceptron | |
| 6 | Deep CNN - Convolutional Neural Network - 2 Filter Stages | |
| 7 | Deeper CNN - Convolutional Neural Network - 4 Filter Stages | |
| 8 | RNN - Recurrent Neural Network | |
| 9 | DNN - Deep Neural Network | |
| 10 | RF - Random Forest | 1.0000 |

## 4.2- Second Stage Results and Discussions for analyzing MNIST

In this stage, we introduce MNIST data (unstructured data set) as one additional complexity over the first stage (structured data set) to explore neural network performance in such an environment. The computation time and accuracy results for three NN algorithms, Deep CNN, RN, and DNN of Fig. 5, show that Deep CNN delivers the highest accuracy but the longer computation time. At the same time, DNN gives the shortest computation time and better accuracy than RNN. Therefore, we exclude RNN from moving to the subsequent stages.

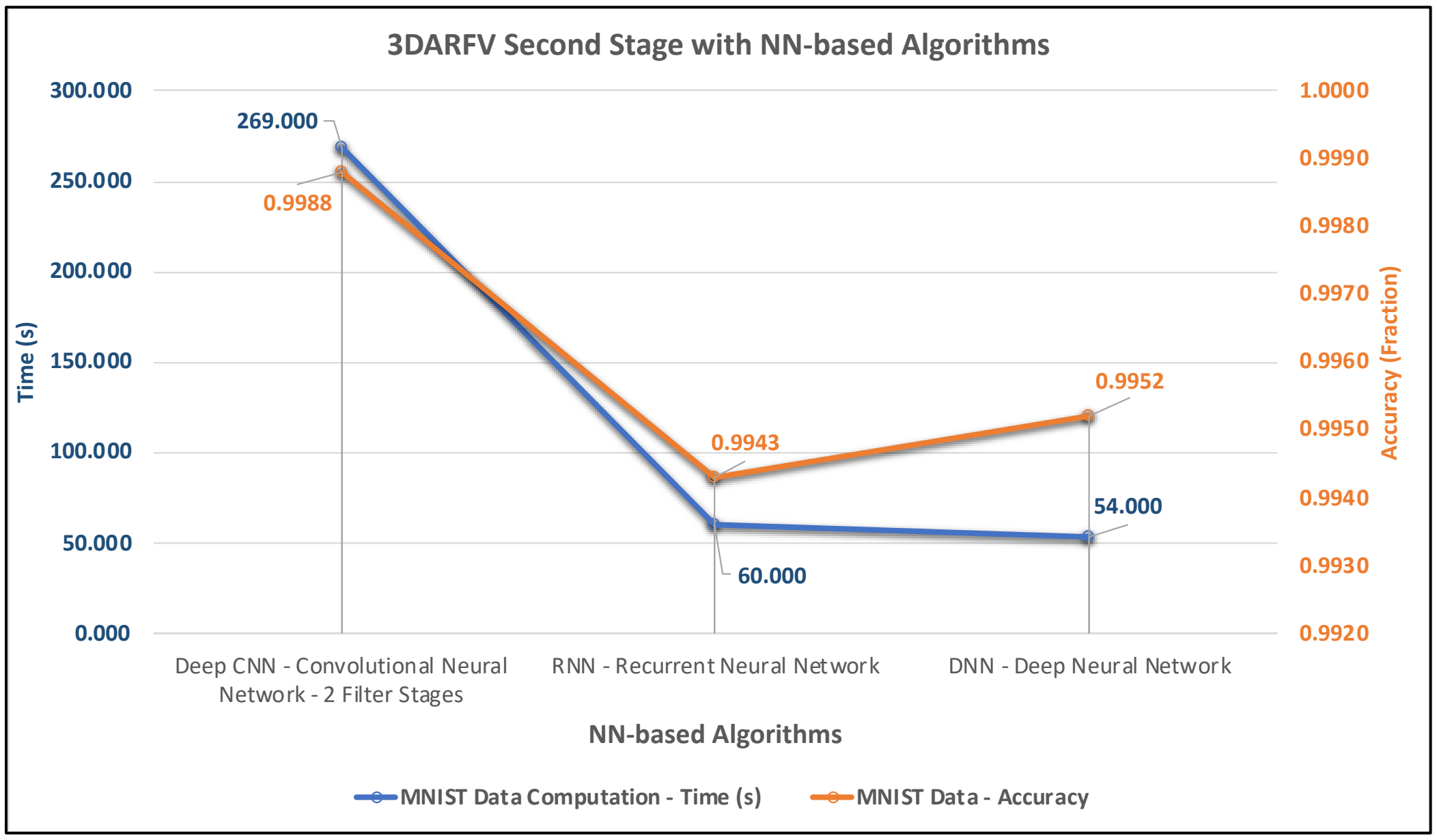

Fig 5. The Second Stage NN-based algorithm's comparison of Accuracy and Computation Time.

## 4.3- Third Stage Results and Discussions for analyzing CIFAR-10

In this stage, we introduce CIFAR-10, a data set that adds another complexity over the second stage by having color and object in the images. In Table 6, we display the results of Deep CNN (two filter stages) and Deeper CNN (four filter stages). The deeper CNN delivered higher accuracy, but at the cost of computation time. In our natural rock of Fig. 3, we have fossils that belong to the same family but of different fossil sizes. Therefore, using CNN is vital for our research in defining the different types of fossils. But Fig. 6 informs us that we need to reduce the amount of CNN computation tasks to have a shorter CNN computation time. Therefore, we recommend performing semantic segmentation run using DNN or RF, separating the fossils segment, and the CNN object detection follows.

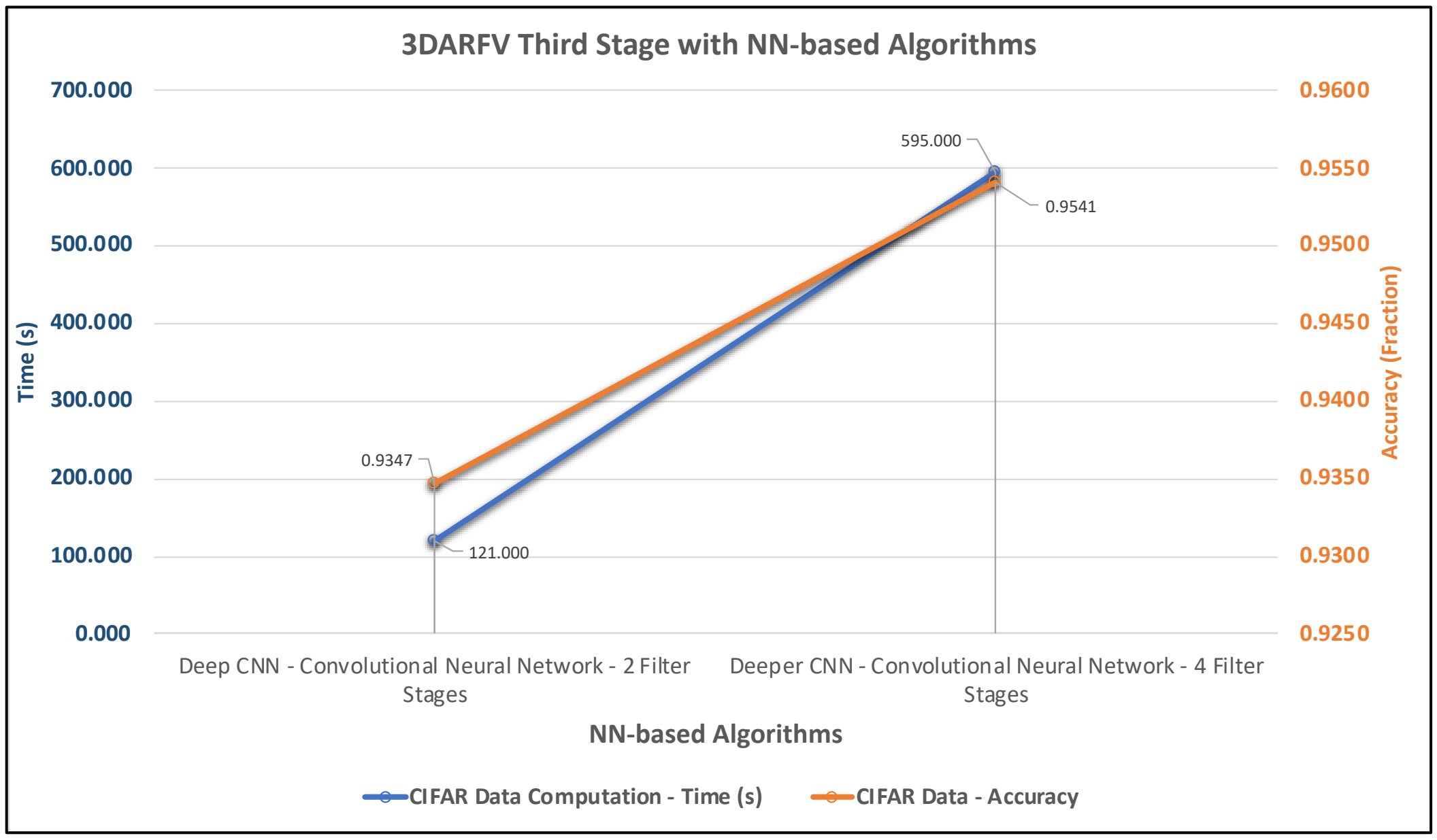

Fig 6. The Third Stage CNN-based algorithm's comparison of Accuracy and Computation Time.

## 4.4- Forth Stage Results and Discussions for analyzing Natural Rock

In this stage, we introduce Savonnières carbonate (SAV-II) rock μCT images, a data set of geoscientist interest to analyze for characterizing rock physical and chemical properties. In Fig. 7, we display the results of SAV-II semantic segmentation, which provides new insight into this heterogeneous fabric of carbonate rocks. The segmentation of PorThN determines the rock's physical and chemical properties [25, 45]: lithology, porosity, permeability, capillary pressure, lithofacies, and relative permeability [46]. We display the comparison between the three finalist algorithms NB, DNN, and RF in Fig. 8, which conclude the RF's ability to deliver the most efficient computation at the utmost accuracy. NB performance shows a shorter computation time than DNN, but DNN is the second-best algorithm because DNN provides the utmost accuracy that NB does not for SAV-II. RF and DNN deliver the best accuracy for optimal vision-data-driven semantic segmentation-based decision-making in heterogeneous fabric. However, RF performs with much higher efficiency than DNN by ~170 folds, as shown in Fig. 8.

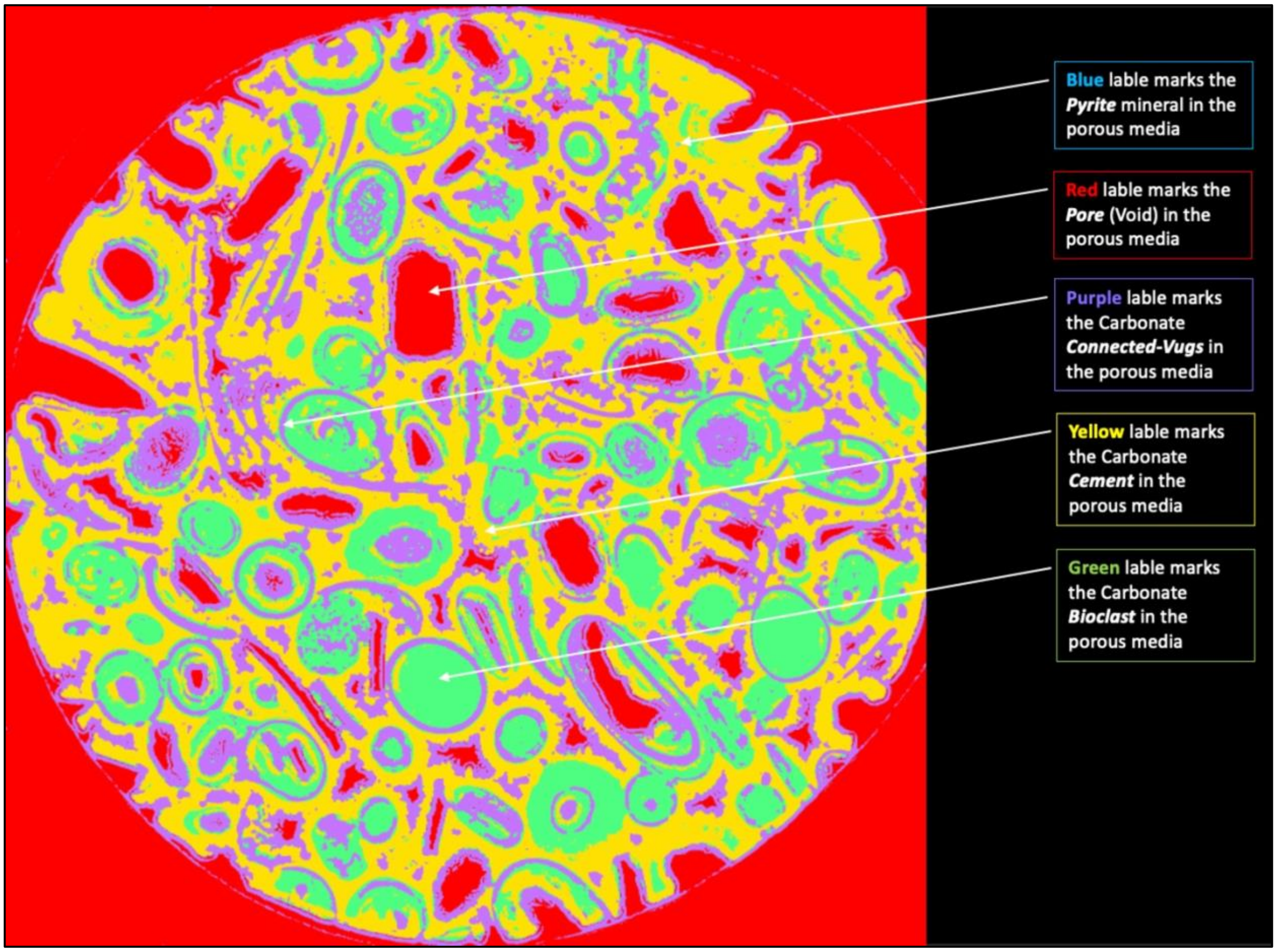

Fig. 7. Semantically Segmented Natural Carbonate Rock X-Ray Micro Computerized Tomography (μCT) Image using 3DARFV.

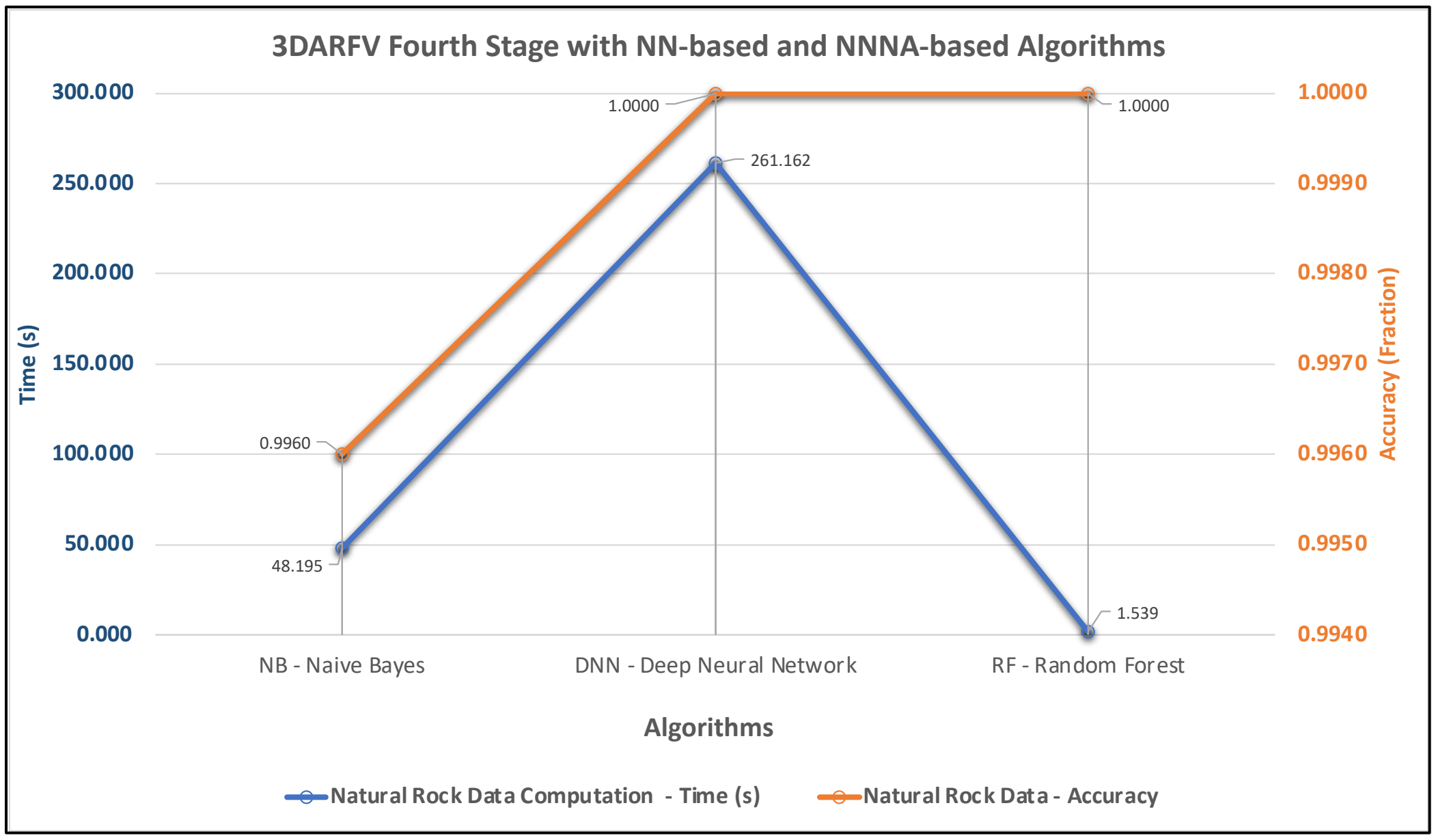

Fig. 8. The Fourth Stage NN and NNNA Algorithm's Comparison for Natural Carbonate Rock X-Ray Micro Computerized Tomography (μCT) Image.

Finally, after establishing the stage-based elimination of algorithms, we measure the time needed for processing 50 images of SAV-II rock μCT 3D stack, to be ~100 seconds (equivalent to 1.5 minutes). While if we consider performing the same semantic segmentation task for the 50 images of SAV-II rock μCT 3D stack with DNN, then we estimate the time to be >13000 seconds (equivalent to 216 minutes which is ~3.5 hours). Therefore, we conclude that RF is the winner. In Fig. 9, we display a sample of 20 semantically segmented SAV-II μCT using 3DARFV.

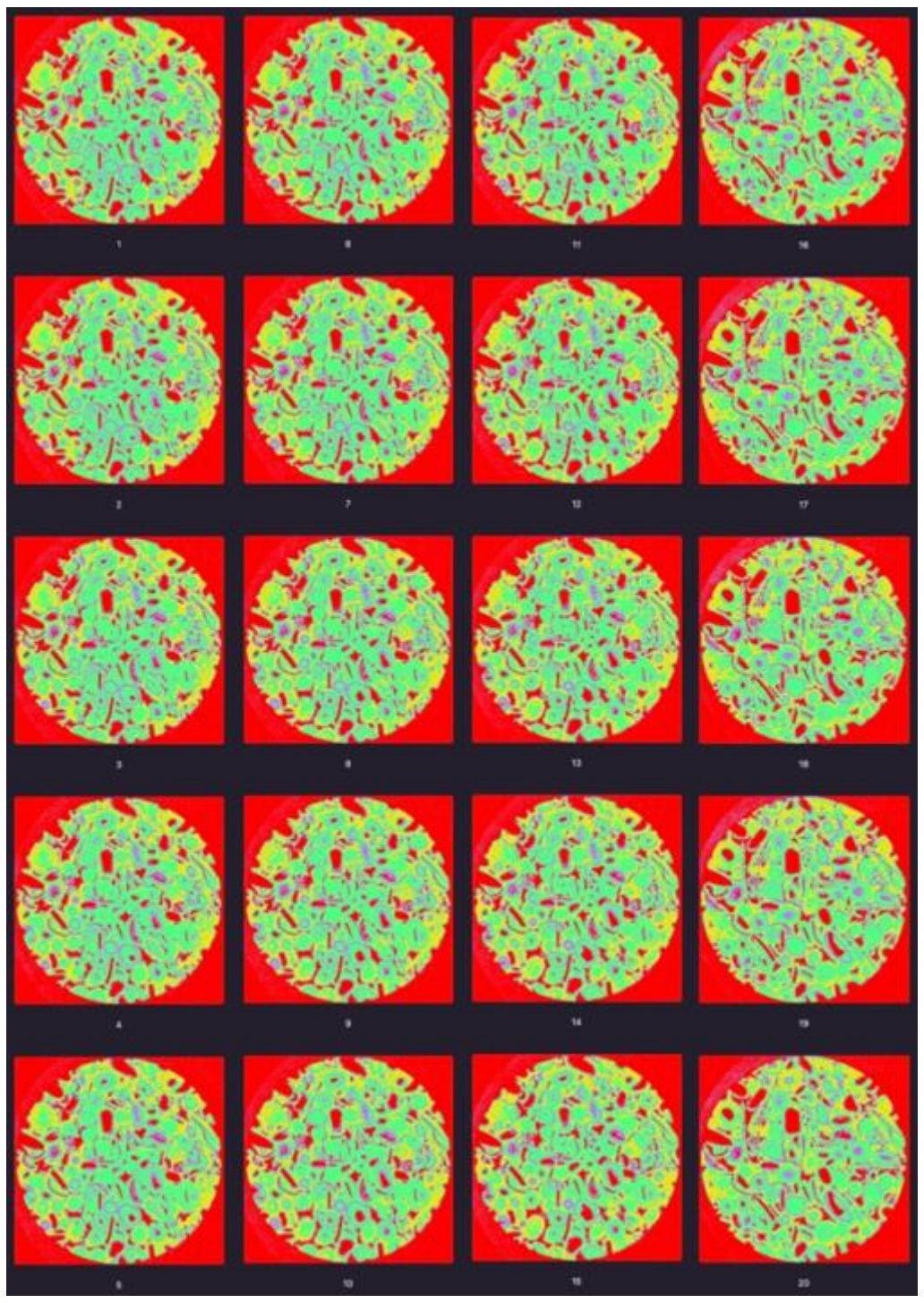

Fig. 9. Natural Carbonate Rock μCT 3D Stack of 20 Semantically Segmented Images using 3DARFV.

## 5. Conclusions and Recommendations

We conclude that any algorithm does not consistently rank the best accuracy for all the different data sets. It is the same regarding algorithm efficiency, where any algorithm does not rate the most efficient for all the other data sets. When accuracy and efficiency jointly form the comparison criteria, an algorithm does not necessarily rank the best for all different data sets. We found that deep learning algorithms generally deliver higher accuracy than NNNA algorithms, except for Random Forest, which showed an utmost accuracy like deep learning. Our experiments showed that NN algorithms have a longer computation time than NNNA. And convolutional neural network delivers higher accuracy when increasing the filters stages, at the price of longer computation time. For 3D vision analysis of a complex combination of multiple objects and semantic segmentation classes, like heterogeneous carbonate rock, we recommend using a convolutional neural network after performing efficient 3DARFD semantic segmentation to reduce CNN computation time. Finally, analyzing 3D images for semantic segmentation with limited computation power, in the case of real-time analysis and remote surveillance operations on Earth, Moon, and Mars, the Random Forest algorithm delivers higher efficiency than Deep Learning algorithms at the utmost accuracy.

## 6. Future Work

To further improve NN algorithms, we encourage investigating the parameters and hyperparameters of the DNN algorithm using AutoML. We also consider studying further SVM accuracy compared to RF because SVM showed higher efficiency than RF. However, the SVM accuracy needs further research to confirm. We also consider fine-tuning the RF parameters to achieve higher efficiency at utmost accuracy, 1.00. Finally, fully automated machine learning algorithm development that understands the machine learning purpose from the data set is a vital research area to reduce the time of testing, comparing, and selecting the suitable algorithm for various applications autonomously.

### Remarks

- This paper is a preprint version for e-print server submission.
- This version was last updated on Feb 20th, 2021.

### Authors Affiliation

- Omar Alfarisi – Advisor Data Architecture at the Department of Information Management at the ADNOC. Guest Lecturer at the Department of Earth Science at the Khalifa University. Adjunct Professor at the Department of Petroleum Engineering at the China University of Petroleum.
- Zeyar Aung – Associate Professor at the Department of Electrical Engineering and Computer Science at the Khalifa University.
- Qingfeng Huang – Senior Reservoir Engineer at the Department of Field Development at the Dragon Oil.
- Ashraf Al-Khateeb – Assistant Professor at the Department of Aerospace Engineering at the Khalifa University.
- Hamed Alhashmi – Team Leader Geoscience at the Department of Field Development at the ADNOC Offshore.
- Mohamed Abdelsalam – Team Leader Petroleum Engineering at the Department of Field Development at the ADNOC Offshore.
- Salem Alzaabi – Manager Field Development at the Department of Field Development at the ADNOC Offshore.
- Haifa Alyazeedi – Manager Information Management at the Department of Group Upstream Information Management at the ADNOC.
- Anthony Tzes – Professor at the Department of Electrical and Computer Engineering at the New York University Abu Dhabi.

### Acknowledgment

The authors thank the support and encouragement received from ADNOC, ADNOC Offshore, Khalifa University, and New York University Abu Dhabi. We express our appreciation to Mr. Yasser Al-Mazrouei, Mr. Ahmed Al-Suwaidi, Mr. Ahmed Al-Hendi, Mr. Ahmed Al-Riyami, Mr. Andreas Scheed, Mr. Hamdan Al-Hammadi, Mr. Khalil Ibrahim.